# Global, Unified Representation of Heterogenous Robot Dynamics Using Composition Operators: A Koopman Direct Encoding Method

H. Harry Asada, *Fellow, IEEE, ASME*

*Abstract*— The dynamic complexity of robots and mechatronic systems often pertains to the hybrid nature of dynamics, where governing equations consist of heterogenous equations that are switched depending on the state of the system. Legged robots and manipulator robots experience contact-noncontact discrete transitions, causing switching of governing equations. Analysis of these systems have been a challenge due to the lack of a global, unified model that is amenable to analysis of the global behaviors. Composition operator theory has the potential to provide a global, unified representation by converting them to linear dynamical systems in a lifted space. The current work presents a method for encoding nonlinear heterogenous dynamics into a high dimensional space of observables in the form of Koopman operator. First, a new formula is established for representing the Koopman operator in a Hilbert space by using inner products of observable functions and their composition with the governing state transition function. This formula, called Direct Encoding, allows for converting a class of heterogenous systems directly to a global, unified linear model. Unlike prevalent data-driven methods, where results can vary depending on numerical data, the proposed method is globally valid, not requiring numerical simulation of the original dynamics. A simple example validates the theoretical results, and the method is applied to a multi-cable suspension system.

## I. INTRODUCTION

ROBOTS perform complex dynamic tasks, interacting with the environment. As a legged robot interacts with a floor, its dynamics change discontinuously [1]. As a manipulator robot manipulates an object, its fingers and arm are subject to non-holonomic geometric constraints, which may dynamically change [2]. These robot dynamics are not unimodal; they are multi-faceted. Overall, their governing dynamics are represented as a combination of heterogenous equations of motion. When an independent set of generalized coordinates are used, the equations of motion are segmented into multiple regions, each representing specific dynamics subject to specific constraints in that region. Depending on contact conditions, constraints, and state locations, a different set of dynamic equations must be applied. This heterogeneous nature of governing equations poses a fundamental challenge for both dynamic modeling and control. The lack of a unified representation that is valid globally for all diverse state locations hinders theoretical analysis of global behaviors.

In system dynamics and control literature, Lyapunov functions have been used for guaranteeing the global stability of a particular class of hybrid systems. Methods have been developed for searching for or learning appropriate Lyapunov functions for given hybrid systems [3], [4]. However, it remains a challenge to find a Lyapunov function for a general class of hybrid systems, especially robotics problems. Poincare's map has been applied to examine convergence and stability of periodic orbits, as observed in legged robots and others [5]. This method requires computation of a series of points on a transversal section, which entails numerical simulations of the original trajectories. Furthermore, this method requires stability analysis of the resultant nonlinear system. Reachability has been studied extensively, and rigorous conditions for reachability have been established [6], [7]. Although reachability is a fundamental property, more detailed analytic tools are required in robotics. Theoretical methodologies for hybrid systems are still limited and analysis is largely dependent on extensive simulations and numerical computations. A global, unified representation of a robot's governing equations could open the door to an alternative approach, which would supplement the existing theories and methodology.

The goal of the current work is to establish a methodology for obtaining a global, uniform representation of multi-faceted, heterogeneous robot dynamics that is amenable to analysis. We aim to construct a global, unified representation directly from the governing equations of heterogeneous robot dynamics. It is uniform in that a single model subsumes all the cases. No segmentation and switching are required. Such a unified representation facilitates the analysis of global behaviors, including stability and convergence.

Our approach utilizes a lifting of the dynamics using supernumerary state variables [8]. Unlike the standard point-wise linearization, this lifting linearization has the potential to construct a global linear model. As illustrated in Fig.1, governing equations in the space of independent state variables

This material is based upon work supported by the National Science Foundation under Grant No. NSF-CMMI 2021625.

H. Harry Asada is with Department of Mechanical Engineering, Massachusetts Institute of Technology, Cambridge, MA 02139 USA (e-mail: asada@mit.edu).

are nonlinear, heterogenous, and are segmented into local regions. As the system representation is lifted to a high-dimensional space, it can be shown that the system behaves linearly and, more importantly, behaves uniformly with no segmentation nor switching. We exploit this linearity of representation for obtaining a unified representation. Although the governing equations are nonlinear and segmented in the original state space, they can be unified with a linear representation in a lifted space. All the heterogenous dynamics, including nonholonomic constraints and discontinuous state transitions among them, may be embedded into a single high-dimensional linear equation.

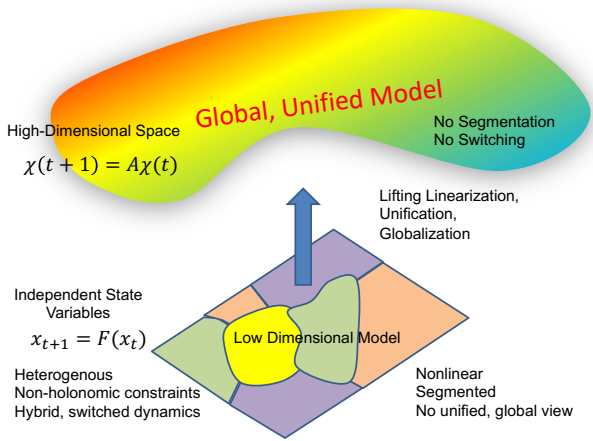

Fig. 1 Global, unified representation of heterogenous dynamics

The theoretical foundation of lifting dynamic modeling was established in the era of Littlewood [9] and Koopman [10] in the 1920's and 30's. Their results showed that a real, single valued function $g: X \to R$, called an observable, which is compositional with a self-map $F: X \to X$, denoted $g \circ F$, can be written as a linear transformation of the observable, $g \circ F = Kg$. This linearity holds although the map $F$ is a nonlinear function. Based on this, a nonlinear, autonomous system, $x_{t+1} = F(x_t)$ and its continuous time form, can be represented as a linear state equation in an infinite-dimensional space. This had a significant impact upon quantum mechanics in the 1930's and 40's. However, no practical application in the engineering field had been found until Igor Mezic pioneered an effective method for applying the Koopman operator to various fields of engineering problems three-quarters of century after the foundational work of Littlewood and Koopman [11].

Mezic and his peers have established the method based on spectral properties of the Koopman operator [12], [13]. The Koopman operator theory has been integrated with Dynamic Mode Decomposition (DMD) and data-driven methods [14], [15]. While the Koopman operator theory underpins the algorithm of DMD from the function analysis viewpoint [16], DMD provides the Koopman method with a practical data-driven technique. Based on the Koopman method, DMD has also been extended to a more general, effective algorithm. Furthermore, the Koopman-DMD approach has been extended to systems with control, or non-autonomous systems [17]. It has opened the door to control system design, in particular, Model Predictive Control (MPC) [18].

Many successful applications of the Koopman-DMD approach have been reported. These include fluid mechanics [13], power systems [12], optimal control [19], and computer vision [20], [21]. In robotics, too, the Koopman-DMD approach has made significant contributions in the last several years. Among others, Abraham and Murphy applied it to active learning for linearizing complex nonlinear dynamics [22]. The method has also been applied to the modeling and control of soft robotic systems, where highly nonlinear, large deformations of soft materials have been modeled accurately [23], [24] and modeling of human-machine systems [25].

It must be noted, however, that a few major drawbacks and limitations exist to the DMD approach. First, it is difficult to guarantee global validity of a model, unless a "compact" set of data is available. Similar to other data-driven approaches, including machine learning, the results are dependent on data used for tuning the system. It is a challenge to collect all the data that guarantee the complete global behaviors of the system. Poor performance may result from insufficient data or from an inappropriate choice of observables. Thus, it is difficult to determine the root cause of performance issues, be it data collection or observable choice.

Another major limitation to the DMD approach is that the algorithm simply assumes that a linear dynamic relationship exists in a high-dimensional space [26]. It does not address whether linearity exists or not. It is questionable particularly when the method is applied to heterogenous dynamical systems that possess discontinuities in state transition and switching of governing equations.

The method of this paper is an alternative to the DMD and other data-driven approaches based on DMD. A linear representation that is valid globally will be obtained directly from governing nonlinear dynamics, which are assumed available. In the following, the composition of an observable with a nonlinear state transition function $F$ will be represented as a linear transformation of the observable function by using an integral kernel. A new formula called Direct Encoding will be developed to obtain the linear state transition matrix of the nonlinear system directly from inner products of observables and their composition with the nonlinear function $F$. Conditions for applying the direct encoding formula to heterogenous nonlinear systems with discontinuity will be discussed.

Most prior works assume that the state transition map $F: X \to X$ is smooth and continuous. This assumption is valid in many applications previously dealt with. However, the state transition map $F: X \to X$ is not always smooth nor continuous when dealing with complex heterogenous dynamic equations. The goal of the current work is to establish a methodology for building a global and unified model of complex dynamical systems, which may include switching among diverse sets of dynamical equations and discontinuous transitions of state.

To this end, properties of the state transition map $F: X \to X$ and a system of observable functions will be examined for finding conditions under which a linear state transition equation exists. Three major results will be obtained; a) a proper choice of observables matched with a given nonlinear state transition function $F$ is required for obtaining a linear dynamic equation in a lifted space; b) if the conditions are met, a global linear model can be constructed through inner product computations over the state space; and c) although the conditions are not satisfied for a class of heterogenous systems, a finite-order model can be used to approximate the heterogenous dynamics under a specific mild condition. Because the method does not use simulation data for training a linear model, it does not possess the drawbacks that the DMD approach and data-driven methods suffer from. The method will be applied to a univariate hybrid dynamical system and a multi-cable heterogenous dynamical system. The implication of the proposed method and remaining open questions will be discussed at the end.

## II. Linear Representation of Nonlinear Dynamics

This section will present the basic formulation of the problem, including a minimum background of functional analysis and composition operators [27]. Although the Koopman Operator theory has been established in a general Banach space [28], Direct Encoding entails inner products. As such, the following formulation assumes that all functions are involved in a Hilbert space. This results in a simple, straightforward transformation of nonlinear dynamics into lifted linear state equations using basic function analysis. This formulation based on inner products leads to the Direct Encoding formula and lays grounds for addressing the applicability of the method to heterogenous dynamical systems in the subsequent section.

### A. Representation of a Compositional Operator Using a Kernel

The fundamental linearity given by the Koopman operator theory can be manifested with use of an integral kernel. Consider a discrete-time, autonomous dynamical system:

$$x_{t+1} = F(x_t) \quad (1)$$
$$y_t = g(x_t) \quad (2)$$

where $x_t \in X \subset \mathcal{R}^n$ is an independent state vector of dimension $n$, $F(x_t)$ is a real-valued vector function $F: X \to X$, which is a self-map called a state transition function, $y_t$ is output, and $g(x_t)$ is an output function. In Koopman operator theory, $g(x_t)$ is called an "observable". Combining (1) and (2) yields

$$g(x_{t+1}) = g[F(x_t)] \quad (3)$$

In this section we aim to examine the properties of $g[F(x_t)]$ from a function analysis viewpoint. In doing so we treat $g[F(x_t)]$ as a real-valued function on $X \subset \mathcal{R}^n$, and deal with a scalar function. In the following, we drop the subscript $t$ to emphasize transformation of the entire function over $X \subset \mathcal{R}^n$, and de-emphasize time evolution. The connection between function transformation and time evolution will be established in the succeeding sub-section.

Consider a Hilbert space $\mathcal{H}$ with an inner product $\langle -, - \rangle: \mathcal{H} \times \mathcal{H} \to C$, expressed as

$$\langle f_1, f_2 \rangle = \int_X f_1(x)\overline{f_2}(x)dx, \quad (4)$$

where $f_1, f_2 \in \mathcal{H}$, and $\overline{f_2}$ represents the complex conjugate of $f_2$. The function in (3) is a composition of function $g$ with a self-map $F$, which is represented using a composition operator as

$$(g \circ F)(x) = g[F(x)] \quad (5)$$

In the case of heterogenous dynamical systems, the state transition function $F$ may be segmented into multiple regions, called "state locations", and the governing equations must be changed depending on the state location. Despite such complexities, there exists a linear relationship between observable function $g$ and its composition with $F$ - that is, $g \circ F$ - as shown next using a kernel.

**Proposition 1**

Let $\mathcal{H}$ be a Hilbert space on $X \subset \mathcal{R}^n$, and let $\{\varphi_1, \varphi_2, \varphi_3, \cdots\}$ be an orthonormal set of basis functions spanning $\mathcal{H}$. Let $F$ be a state transition function $F: X \to X$ and $g$ be a function in the Hilbert space.

$$g \in \mathcal{H} \quad (6)$$

Then, there exists a linear relationship between function $g$ and its composition with $F$,

$$(g \circ F)(x) = \int_X \kappa(x, \xi)g(\xi)d\xi \quad (7)$$

where $\xi$ is an integral variable, and $\kappa: X \times X \to C$ is a kernel given by

$$\kappa(x, \xi) = \sum_{k=1}^{\infty} \varphi_k[F(x)]\overline{\varphi}_k(\xi). \quad (8)$$

***Proof***

From (6), $g$ can be expanded in the orthonormal basis $\{\varphi_1, \varphi_2, \varphi_3, \cdots\}$,

$$g = \sum_{k=1}^{\infty} \langle g, \varphi_k \rangle \varphi_k \quad (9)$$

By using this expansion, the compositional function $g \circ F$ can be written as

$$g \circ F = \left(\sum_{k=1}^{\infty} \langle g, \varphi_k \rangle \varphi_k\right) \circ F \quad (10)$$

Distributing the compositional operator yields

$$g \circ F = \sum_{k=1}^{\infty} \langle g, \varphi_k \rangle (\varphi_k \circ F) \quad (11)$$

From the definition of inner product (4)

$$(g \circ F)(x) = \sum_{k=1}^{\infty} \int_X g(\xi)\overline{\varphi}_k(\xi)d\xi \, (\varphi_k \circ F)(x)$$
$$= \int_X \sum_{k=1}^{\infty} \varphi_k[F(x)]\overline{\varphi}_k(\xi)g(\xi)d\xi = \int_X \kappa(x, \xi) g(\xi)d\xi \quad (12)$$

Q.E.D.

Note that the kernel $\kappa(x, \xi)$ encodes the state transition function $F(x)$ with the basis functions $\{\varphi_1, \varphi_2, \varphi_3, \cdots\}$. The kernel does not depend on the function $g \in \mathcal{H}$, and it can be applied to arbitrary functions in $\mathcal{H}$:

$$(g_i \circ F)(x) = \int_X \kappa(x, \xi)g_i(\xi)d\xi, \quad g_i \in \mathcal{H}, i \in \mathbb{N} \quad (13)$$

Concatenating these in an infinite dimensional vector of functions $g_i \circ F$ yields

$$\begin{bmatrix} (g_1 \circ F)(x) \\ (g_2 \circ F)(x) \\ \vdots \end{bmatrix} = \int_X \kappa(x,\xi) \begin{bmatrix} g_1(\xi) \\ g_2(\xi) \\ \vdots \end{bmatrix} d\xi \qquad (14)$$

These observables $g_1, g_2, \cdots$ collectively represent the state of the system in a lifted space of infinite dimension. Denoting the concatenated observables by

$$\chi(x) \triangleq \begin{bmatrix} g_1(x) \\ g_2(x) \\ \vdots \end{bmatrix}, \qquad (15)$$

we aim to obtain a linear state transition equation. Under certain conditions, such a linear state transition equation exists, as discussed next.

*B. Linear State Transition*

While the kernel representation given by eq.(7) manifests the linear transformation of observables to their compositions with the nonlinear state transition function $F$, it is not in the form of linear state equation, which we need in engineering applications. The following Proposition provides the conditions for the existence of a linear state equation.

**Proposition 2**

Let $\{\varphi_1, \varphi_2, \varphi_3, \cdots\}$ be an orthonormal set of basis functions in a Hilbert space $\mathcal{H}$ on $X \subset \mathcal{R}^n$, and let $F$ be a state transition function $F: X \to X$. If the compositional function $\varphi_i \circ F$ is in the Hilbert space,

$$\varphi_i \circ F \in \mathcal{H}, \qquad i \in \mathbb{N} \qquad (16)$$

then there exists a linear state transition matrix $A$ of infinite dimension that transfers a lifted state in the form of $\chi(x) = [g_1(x) \quad g_2(x) \quad \cdots]^T$, where $\{g_1, g_2, g_3, \cdots\}$ forms an orthonormal basis in $\mathcal{H}$, to the lifted state associated to the state transition function $F$,

$$\chi[F(x)] = A \chi(x) \qquad (17)$$

where $\chi[F(x)] = [g_1[F(x)] \quad g_2[F(x)] \quad \cdots]^T$, and the state transition matrix $A$ is given by

$$A = \sum_{j=1}^\infty \begin{bmatrix} \langle g_1, \varphi_j \rangle \\ \langle g_2, \varphi_j \rangle \\ \vdots \end{bmatrix} [\langle \varphi_j \circ F, g_1 \rangle \quad \langle \varphi_j \circ F, g_2 \rangle \quad \cdots] \quad (18)$$

***Proof***

From (11) in the proof of Proposition 1,

$$\begin{bmatrix} g_1 \circ F \\ g_2 \circ F \\ \vdots \end{bmatrix} = \begin{bmatrix} \sum_j \langle g_1, \varphi_j \rangle \varphi_j \circ F \\ \sum_j \langle g_2, \varphi_j \rangle \varphi_j \circ F \\ \vdots \end{bmatrix}$$

$$= \sum_{j=1}^\infty \begin{bmatrix} \langle g_1, \varphi_j \rangle \\ \langle g_2, \varphi_j \rangle \\ \vdots \end{bmatrix} \varphi_j \circ F \qquad (19)$$

Because $\varphi_i \circ F \in \mathcal{H}$ and $\{g_1, g_2, g_3, \cdots\}$ is an orthonormal set of basis functions in $\mathcal{H}$, function $\varphi_i \circ F$ can be expanded in the basis $\{g_1, g_2, g_3, \cdots\}$:

$$\varphi_j \circ F = \sum_{k=1}^\infty \langle \varphi_j \circ F, g_k \rangle g_k \qquad (20)$$

Substituting this into (19) yields

$$\chi[F(x)] = \begin{bmatrix} g_1[F(x)] \\ g_2[F(x)] \\ \vdots \end{bmatrix}$$

$$= \sum_{j=1}^\infty \begin{bmatrix} \langle g_1, \varphi_j \rangle \\ \langle g_2, \varphi_j \rangle \\ \vdots \end{bmatrix} \sum_{k=1}^\infty \langle \varphi_j \circ F, g_k \rangle g_k(x) =$$

$$\sum_{j=1}^\infty \begin{bmatrix} \langle g_1, \varphi_j \rangle \\ \langle g_2, \varphi_j \rangle \\ \vdots \end{bmatrix} [\langle \varphi_j \circ F, g_1 \rangle, \langle \varphi_j \circ F, g_2 \rangle, \cdots] \begin{bmatrix} g_1(x) \\ g_2(x) \\ \vdots \end{bmatrix}$$

$$= A \chi(x) \qquad (21)$$

Thus, the Proposition has been proven. Q.E.D.

A special case of Proposition 2 is to use $\{\varphi_1, \varphi_2, \varphi_3, \cdots\}$ for the second set of orthonormal basis functions $\{g_1, g_2, g_3, \cdots\}$. In this case, the matrix $A$ is simplified.

**Corollary 1**

In Proposition 2, if $g_i = \varphi_i$, $i = 1, 2, 3, \cdots$, the state transition matrix $A$ is simplified to

$$\bar{A} = \begin{bmatrix} \langle \varphi_1 \circ F, \varphi_1 \rangle & \langle \varphi_1 \circ F, \varphi_2 \rangle & \cdots \\ \langle \varphi_2 \circ F, \varphi_1 \rangle & \langle \varphi_2 \circ F, \varphi_2 \rangle & \cdots \\ \vdots & \vdots & \ddots \end{bmatrix} \qquad (22)$$

This can be shown directly by replacing $\{g_1, g_2, g_3, \cdots\}$ by $\{\varphi_1, \varphi_2, \varphi_3, \cdots\}$ in (18).

In Proposition 2, the observables $\{g_1, g_2, g_3, \cdots\}$ must be an orthonormal set of basis functions. While this condition is restrictive, it can be relaxed. Let $\{g_1, g_2, g_3, \cdots\}$ be an independent and complete set of observables spanning the Hilbert space. Using the orthonormal basis functions, each observable can be expanded to $g_i = \sum \langle g_i, \varphi_k \rangle \varphi_k$, or collectively expressed as

$$\begin{bmatrix} g_1 \\ g_2 \\ \vdots \end{bmatrix} = C \begin{bmatrix} \varphi_1 \\ \varphi_2 \\ \vdots \end{bmatrix} \qquad (23)$$

where

$$C = \begin{bmatrix} \langle g_1, \varphi_1 \rangle & \langle g_1, \varphi_2 \rangle & \cdots \\ \langle g_2, \varphi_1 \rangle & \langle g_2, \varphi_2 \rangle & \cdots \\ \vdots & \vdots & \ddots \end{bmatrix} \qquad (24)$$

Because both $\{g_1, g_2, g_3, \cdots\}$ and $(\varphi_1, \varphi_2, \varphi_3, \cdots)$ are independent and complete basis functions, the matrix $C$ is non-singular. Thus, the following corollary allows us to use merely independent basis functions.

**Corollary 2**

In Proposition 2, if $\{g_1, g_2, g_3, \cdots\}$ are an independent and complete set of basis functions, the state transition matrix $A$ in (17) is given by

$$A = C\bar{A}C^{-1} \qquad (25)$$

where $\bar{A}$ has been given by (22) and $C$ by (24).

III. MAIN RESULTS

This section presents a new formulation of the Koopman operator through the use of inner products. The linear representation is globally valid and does not depend on data.

Its applicability to heterogenous dynamical systems will be discussed.

*A. The Direct-Encoding Method Using Inner-Products*

Incorporating Proposition 2 and Corollaries 1 and 2 and constructing inner products, now we obtain the following main proposition.

**Proposition 3**

If observables $\{g_1, g_2, g_3, \cdots\}$ form an independent and complete set of basis functions, spanning a Hilbert space $\mathcal{H}$, and their compositions with a state transition function $F: X \rightarrow X$ are involved in the Hilbert space:

$$g_i \circ F \in \mathcal{H}, \quad i = 1, 2, 3, \cdots \quad (26)$$

then the following state transition equation exists

$$\chi[F(x)] = A\,\chi(x) \quad (27)$$

with a lifted state and a state transition matrix given, respectively, by

$$\chi = \begin{bmatrix} g_1 \\ g_2 \\ \vdots \end{bmatrix}, \quad A = QR^{-1} \quad (28)$$

where the matrices $R$ and $Q$ consist of inner products given by

$$R = \begin{bmatrix} \langle g_1, g_1 \rangle & \langle g_1, g_2 \rangle & \cdots \\ \langle g_2, g_1 \rangle & \langle g_2, g_2 \rangle & \cdots \\ \vdots & \vdots & \ddots \end{bmatrix} \text{ and }$$

$$Q = \begin{bmatrix} \langle g_1 \circ F, g_1 \rangle & \langle g_1 \circ F, g_2 \rangle & \cdots \\ \langle g_2 \circ F, g_1 \rangle & \langle g_2 \circ F, g_2 \rangle & \cdots \\ \vdots & \vdots & \ddots \end{bmatrix}. \quad (29)$$

***Proof***

From (23),

$$\begin{bmatrix} \varphi_1(x) \\ \varphi_2(x) \\ \vdots \end{bmatrix} = C^{-1} \begin{bmatrix} g_1(x) \\ g_2(x) \\ \vdots \end{bmatrix} \quad (30)$$

Substituting $F(x)$ into $x$ in (30)

$$\begin{bmatrix} \varphi_1[F(x)] \\ \varphi_2[F(x)] \\ \vdots \end{bmatrix} = C^{-1} \begin{bmatrix} g_1[F(x)] \\ g_2[F(x)] \\ \vdots \end{bmatrix} \quad (31)$$

From the assumption (26), $g_i \circ F \in \mathcal{H}, \forall i$. Therefore, $\varphi_i \circ F \in \mathcal{H}, \forall i$. From Proposition 2, it is confirmed that the linear state transition equation (27) exists.

Post-multiplying a row vector function $[g_1, g_2, g_3, \cdots]$ to (27) and integrate over $X$ yields,

$$\int_X \begin{bmatrix} g_1 \circ F \\ g_2 \circ F \\ \vdots \end{bmatrix} [g_1, g_2, \cdots] dx$$

$$= A \int_X \begin{bmatrix} g_1 \\ g_2 \\ \vdots \end{bmatrix} [g_1, g_2, \cdots] dx \quad (32)$$

where the matrix $A$ is factored out. Each integral in the vector outer product is the inner product of the corresponding two functions. Therefore,

$$\begin{bmatrix} \langle g_1 \circ F, g_1 \rangle & \langle g_1 \circ F, g_2 \rangle & \cdots \\ \langle g_2 \circ F, g_1 \rangle & \langle g_2 \circ F, g_2 \rangle & \cdots \\ \vdots & \vdots & \ddots \end{bmatrix}$$

$$= A \begin{bmatrix} \langle g_1, g_1 \rangle & \langle g_1, g_2 \rangle & \cdots \\ \langle g_2, g_1 \rangle & \langle g_2, g_2 \rangle & \cdots \\ \vdots & \vdots & \ddots \end{bmatrix} \quad (33)$$

or

$$Q = AR \quad (34)$$

Since $\{g_1, g_2, g_3, \cdots\}$ are independent, matrix $R$ is non-singular. Therefore, (28) is obtained. Q.E.D.

Note that computation of $A = QR^{-1}$ does not require $\{\varphi_1, \varphi_2, \cdots\}$ nor $C$ and $C^{-1}$ involved in (25). Once we know the existence of (27), matrix $A$ can be computed directly by taking inner products between $g_i$ and its composition with $F$. The inner products $\langle g_i, g_j \rangle$ and $\langle g_i \circ F, g_j \rangle$ exist and are well-defined because both $g_i$ and $g_j \circ F$ belong to the Hilbert space.

Note that this formula for obtaining a linear system matrix $A$ is an alternative to the prevailing Koopman DMD method, which assumes the existence of a linear system matrix $A$ in data and computes the matrix by Least Squares Estimate. The $A$ matrix based on the Direct Encoding is globally valid, while the DMD and its variants are data dependent.

*B. Weak Existence Conditions for Finite-Dimensional Approximation*

In Proposition 2, it is assumed that the compositional function $\varphi_i \circ F$ is in a Hilbert space for all $i$, including $\infty$. As will be detailed in the succeeding section, this condition is difficult to satisfy when applying the compositional operator method to heterogenous systems, specifically to hybrid systems with discontinuous state transition. This section explores the possibility to apply the compositional operator method to hybrid systems by replacing the strong condition, $\varphi_i \circ F \in \mathcal{H}$, $\forall i \in \mathbb{N}$, by its weaker condition, $\varphi_i \circ F \in \mathcal{H}$, $\forall i < \infty$. Namely, the compositional function $\varphi_i \circ F$ must be in a Hilbert space only for an arbitrary, finite $i$. This implies truncation of the infinite-dimensional lifted space, resulting in approximation of the linear state transition. From Proposition 2 and Corollary 1, we obtain:

**Proposition 4**

Let $\{\varphi_1, \varphi_2, \varphi_3, \cdots\}$ be an orthonormal set of basis functions in a Hilbert space $\mathcal{H}$ on $X \subset \mathcal{R}^n$, and let $F$ be a state transition function $F$. If the first $m$ compositional functions $\varphi_i \circ F$ are in the Hilbert space,

$$\varphi_i \circ F \in \mathcal{H}, \quad 1 \leq \forall i \leq m < \infty \quad (35)$$

then there exists a linear state transition matrix $A_{m\infty}$ that transfers an infinite-dimensional lifted state in the form of $\chi(x) = [\varphi_1(x) \quad \varphi_2(x) \quad \cdots]^T$ to its finite-dimensional state consisting of composite functions.

$$\begin{bmatrix} \varphi_1 \circ F \\ \varphi_2 \circ F \\ \vdots \\ \varphi_m \circ F \end{bmatrix} = A_{m\infty} \begin{bmatrix} \varphi_1 \\ \varphi_2 \\ \vdots \end{bmatrix} \quad (36)$$

where $A_{m\infty}$ is a $m$ by $\infty$ matrix given by

$$A_{m\infty} = \begin{bmatrix} \langle \varphi_1 \circ F, \varphi_1 \rangle & \langle \varphi_1 \circ F, \varphi_2 \rangle & \cdots \\ \langle \varphi_2 \circ F, \varphi_1 \rangle & \langle \varphi_2 \circ F, \varphi_2 \rangle & \cdots \\ \vdots & \vdots & \cdots \\ \langle \varphi_m \circ F, \varphi_1 \rangle & \langle \varphi_m \circ F, \varphi_2 \rangle & \cdots \end{bmatrix} \quad (37)$$

**Proof** Omitted.

Proposition 4 shows that, with a finite number of compositional functions being in the Hilbert space, the one-step time-evolution of the finite number of lifted state variables can be predicted exactly. Since (36) predicts only the finite number of state variables, the prediction of consecutive state transitions may not be exact. Nonetheless, truncated state variables and system matrix may yield reasonable accuracy for a large $m$.

When the orthonormal basis functions $\{\varphi_1, \varphi_2, \varphi_3, \cdots\}$ are replaced by an independent and complete set of basis functions, as in Proposition 3, additional approximation must be made in relating $\{\varphi_1, \varphi_2, \cdots, \varphi_m\}$ to $\{g_1, g_2, \cdots, g_m\}$. Namely, in lieu of Corollary 2, matrices $\bar{A}, C, C^{-1}$ are truncated to $\bar{A}_m \in \mathcal{R}^{m \times m}$, $C_m \in \mathcal{R}^{m \times m}$, and $C_m^{-1} \in \mathcal{R}^{m \times m}$, respectively,

$$A_m = C_m \bar{A}_m C_m^{-1}. \quad (38)$$

where $A_m$ yields only approximate state transition.

## IV. AN ILLUSTRATIVE EXAMPLE

Consider a univariate nonlinear dynamical system on $x \in X = [0,1]$.

$$x_{t+1} = F(x_t), \quad x_t \in [0,1] \quad (39)$$

where

$$F(x) = \begin{cases} cx - cx^*; & 0 \le x \le x^* \\ bx + a - bx^*; & x^* < x \le 1 \end{cases} \quad (40)$$

Figure 2-(a) shows this state transition function with parameters, $a, b, c$, and $x^*$. This system consists of two distinct linear dynamics in two segmented regions or state locations. Note that this piecewise linear system has a discontinuity at $x = x^*$. It is heterogeneous and a type of hybrid system.

From (8) the kernel $\kappa(x, \xi)$ that encodes this state transition equation $F(x)$ is obtained by using the following exponential trigonometric functions, which form orthonormal basis functions spanning a Hilbert space.

$$\varphi_k = \exp(2\pi i k x), \quad k \in \mathbb{Z} \quad (41)$$

where $i$ is imaginary number, and the basis functions are renumbered from $k = 1, 2, 3, \cdots$ to $k = \cdots, -2, -1, 0, 1, 2, \cdots$. The kernel should be able to map an arbitrary observable in the Hilbert space to its composition with $F(x)$. Fig. 2-(b)~(e) confirm this property. All of the observables, $g_1 = x, g_2 = x^2, g_3 = x^3, g_4 = \cos 2\pi x$, are perfectly transformed to their compositional functions with the same kernel. Namely, the right-hand side of (7) computed with the kernel agrees with the true compositional function of the left-hand side.

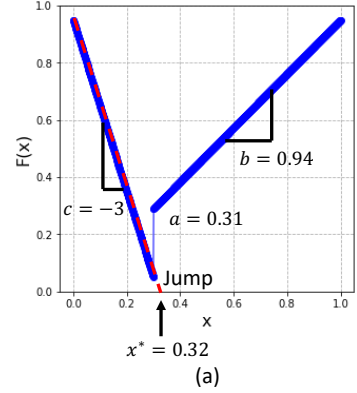

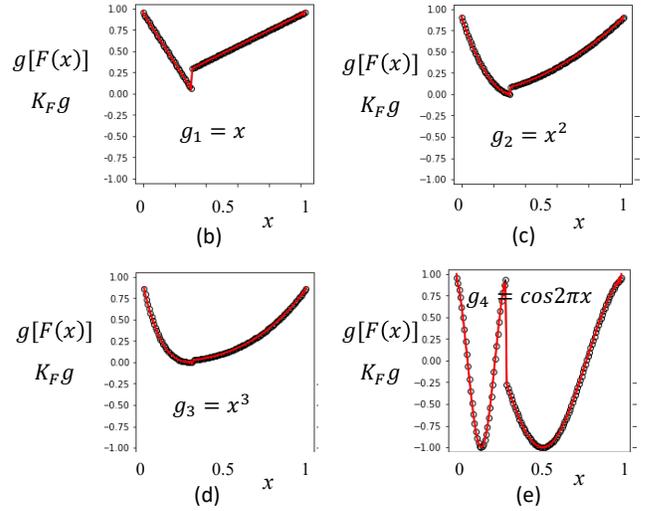

Fig. 2 Example of univariate heterogenous dynamical system (a) The state transition function is heterogenous, consisting of two linear systems combined at x*, where a discontinuous jump occurs. (b)~(e): These plots show that the Koopman operator can transform an arbitrary observable in the Hilbert space to its composition with F(x). The linear transformation shown with black circles can reproduce the correct nonlinear function g[F(x)] shown in red. Four different observables are tested; (b) $g_1 = x$, (c) $g_2 = x^2$, (d) $g_3 = x^3$, and (e) $g_4 = \cos 2\pi x$.

For this kernel to be transformed to the time-evolution, state transition matrix $A$, the conditions of Proposition 2 must be satisfied. In particular,

$$\varphi_k \circ F \in \mathcal{H}, \quad k \in \mathbb{Z} \quad (42)$$

Using the exponential trigonometric function, we obtain:

$$(\varphi_k \circ F)(x) = \exp(2\pi i k F(x)) \quad (43)$$

This compositional function can be expanded in the exponential trigonometric basis for an arbitrary *finite k*. Despite the discontinuity due to the jump in $F(x)$, function $\varphi_k \circ F$ satisfies the following four conditions required for the existence of Fourier series [29]:

1) $\varphi_k \circ F$ is absolutely integrable.
2) There are a finite number of discontinuities within a finite interval.

3) All the discontinuities are finite.
4) $\varphi_k \circ F$ is of bounded variation; there are no more than a finite number of maxima and minima in any finite interval.

With the existence of a Fourier series, $\varphi_k \circ F$ is in a Hilbert space for any finite $k$. This allows us to construct the state-transition matrix based on Proposition 4.

Suppose that the following real-valued observables are selected as an independent set of basis functions, which are complete,

$$g_1 = 1, \ g_{2n} = \cos 2\pi n x, g_{2n+1} = \sin 2\pi n x, \quad n \in \mathbb{N} \tag{44}$$

Conversion of $\varphi_k$ to $g_j$ is given by a nonsingular matrix $C$ and its inverse $C^{-1}$, too, can be computed easily. See Appendix A. These matrices can be truncated to $m \times m$ matrices, and the time-evolution state transition matrix $A_m = C_m \bar{A}_m C_m^{-1}$ is obtained.

This $A_m$ matrix should be able to predict the time-evolution of the lifted state $\chi_t = [g_1, g_2, \cdots, g_m]^T$ mapped to $\chi_{t+1}$. Fig.3 shows the comparison of the state transition based on the linear model, $\chi_{t+1} = A\chi_t$, to the true transition through the original nonlinear function $x_{t+1} = F(x_t)$. Due to the discontinuity in $F(x)$, the true time evolution of the system shows a series of pronounced jumps, as shown by the solid lines in the figures. Nonetheless, the linear model can track this response. For small numbers of observables, truncated at $m = 17$ and 33, the prediction errors are significant. However, as the number of the observables increases, $m = 1025$, the linear model can almost perfectly predict the complex response. Fig. 3-(e) shows mean squared error vs. the number of observables. It is confirmed as the number of observables increases the error approaches zero in the mean. With finite $m$, this is still an approximation. However, the accuracy is high enough for practical applications. Note that the existence conditions hold for any large $m$ that is finite.

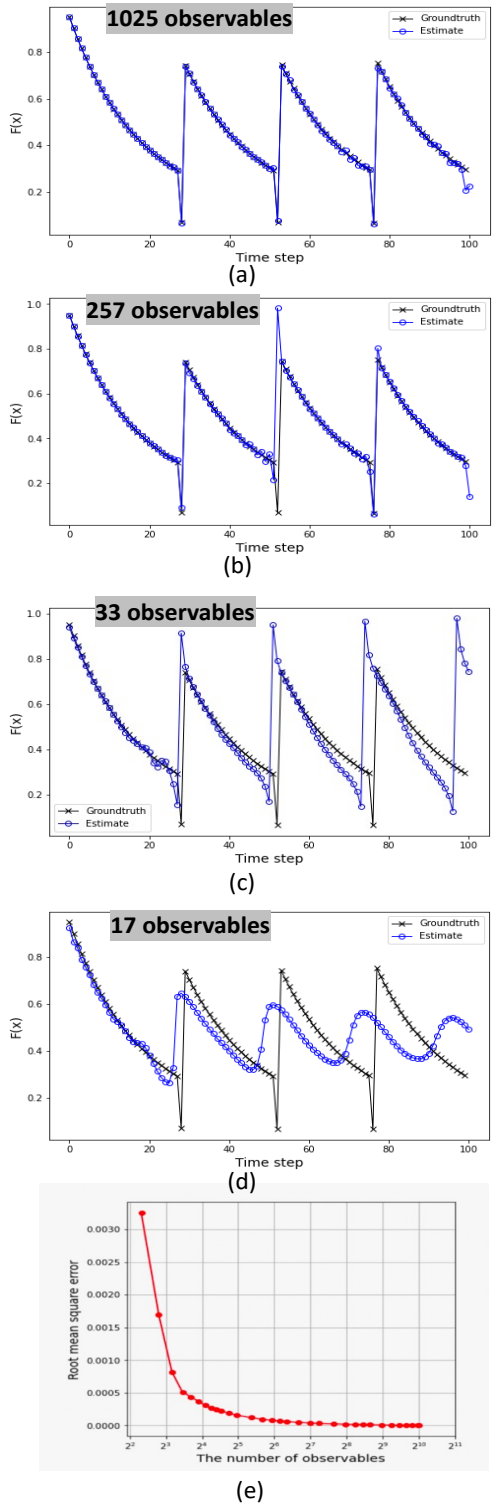

Fig. 3 Evaluation of the linear lifted model in terms of time-evolution prediction accuracy

Black lines in (a)~(d) show the true time trajectory of the hybrid dynamical system, compared to the ones predicted with the linear lifted model shown in blue. (a) With 1,025 observables, almost perfect prediction was achieved, while the prediction accuracy declines with 257 observables (b), 33 observables (c), and 17 observables (d). (e) summarizes the prediction accuracy in terms of root mean square error for different numbers of observables. The error converges to zero in the mean.

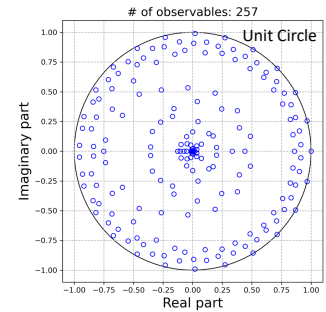

Figure 4 Plot of poles of $A_{257}$ on complex plane

This linear model is global; it is valid for the entire domain. Unlike data-driven methods, where results depend on how data are collected, this model is derived directly from the governing model. This model is also uniform. Despite the discontinuous switching of the two linear models involved in the system, the linear model represents the heterogeneous system behavior with a single system of linear dynamic equations uniformly in the lifted space.

This linear model facilitates dynamic analysis of the hybrid system. Fig.4 shows the plot of poles obtained from the $A_{257}$ matrix. Note that all the poles are on or within the unit circle. Clearly, the system is marginally stable. A group of poles are on the unit circle, indicating that the system has undamped oscillation modes. This agrees with the time trajectories in Fig.3, where the system converges to a periodic orbit.

## V. MULTI-CABLE DYNAMIC MANIPULATION

Now the method is applied to a robotics problem. Contact-noncontact switching of dynamics is a challenging problem but is a fundamentally important issue in broad fields of robotics. In multi-cable manipulation, for example, an object suspended with cables experiences discrete switching of governing equations as each cable alters between slack and taut conditions [30].

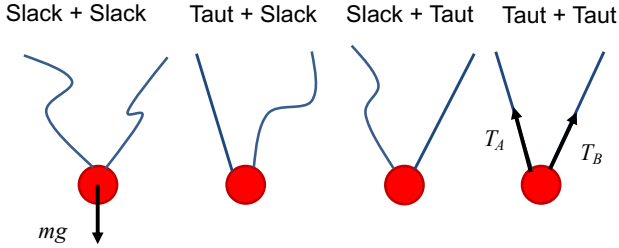

Fig. 5 Diverse states of a point mass suspended with two cables

As shown in Fig.5, a point mass $m$ is connected to a pair of cables. When the point mass is released from a certain height, it drops like a projectile object, but it bounces back when the cable(s) become taut, like a ball bouncing on a floor. Each cable can bear only a unidirectional load; it goes slack when a compressive load acts on the cable. The constraints due to this unidirectional nature of cable suspension are therefore non-holonomic. Depending on which cable is taut or slack, different dynamic equations are switched.

Let $x_m$ and $y_m$ be coordinates of the point mass. Equations of motion are segmented to 4 regions.

$$m\begin{pmatrix}\ddot{x}_m\\\ddot{y}_m\end{pmatrix} = -m\mathbf{g}, \qquad \begin{pmatrix}x_m\\y_m\end{pmatrix} \in D_1$$

$$m\begin{pmatrix}\ddot{x}_m\\\ddot{y}_m\end{pmatrix} = T_A n_A - m\mathbf{g}, \qquad \begin{pmatrix}x_m\\y_m\end{pmatrix} \in D_2$$

$$m\begin{pmatrix}\ddot{x}_m\\\ddot{y}_m\end{pmatrix} = T_B n_B - m\mathbf{g}, \qquad \begin{pmatrix}x_m\\y_m\end{pmatrix} \in D_3 \qquad (45)$$

$$m\begin{pmatrix}\ddot{x}_m\\\ddot{y}_m\end{pmatrix} = T_A n_A + T_B n_B - m\mathbf{g}, \begin{pmatrix}x_m\\y_m\end{pmatrix} \in D_4$$

where $T_A$ and $T_B$ are tensions of cables A and B, respectively, $n_A$ and $n_B$ are unit vectors pointing in the direction of the two cables, and $D_1, \cdots, D_4$ are regions of the point mass locations corresponding to $D_1$: both cables are slack, $D_2$: cable A is taut and cable B is slack, $D_3$: cable B is taut and cable A is slack, and $D_4$: both cables are taut. Both cables have a high extensile stiffness but zero stiffness for a compressive load. The cables are assumed to be massless and to have small damping. The independent state variables are:

$$x = (x_m, y_m, \dot{x}_m, \dot{y}_m)^T \in X \subset \mathcal{R}^4 \qquad (46)$$

The dynamic range of the system is finite, given the length of each cable.

Gaussian Radial Basis Functions (RBFs) are used as observables:

$$g_k(x) = \psi_k(\|x - c_k\|), \qquad k \in \mathbb{N} \qquad (47)$$

where in each RBF $g_k$, $c_k$ is the center point. By construction, RBFs are guaranteed to be independent [31]. The number of RBFs, that is, the order of the lifted space, must be finite for implementing the observables. In total, 1,500 RBFs are used for covering the 4-dimensional state space within the bounded dynamic range. They are placed at optimal locations by clustering sample points using the $k$ means ++ algorithm. More details are described in Appendix B.

The independence of the RBFs can be confirmed by examining whether the $R$ matrix in (29) is non-singular. The $Q$ matrix is determined by computing the inner products between individual RBFs and their composition with the state transition function $F(x)$:

$$\langle g_k \circ F, g_j \rangle = \int_X \psi_k(\|F(x) - c_k\|)\psi_j(\|x - c_j\|)dx. \qquad (48)$$

From these $R$ and $Q$ matrices, the $A$ matrix is obtained: $A = QR^{-1}$.

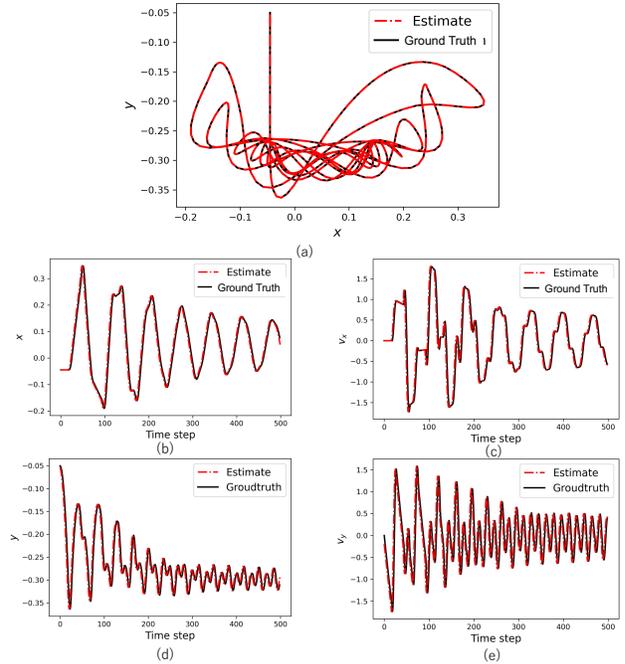

Fig. 6 Accuracy of time evolution prediction for the point mass suspended with two cables. (a) Trajectories in the $xy$-plane. The black solid line shows the trajectory generated with the original nonlinear, heterogenous dynamic equations, while the red dash-and-dot line shows the trajectory predicted with the linear lifted model. The two trajectories overlap, indicating high accuracy of prediction. (b) ~(e) show the time profiles of position $x$ and its velocity and position $y$ and its velocity, respectively.

Fig. 6 shows the comparison between the point mass trajectory of the original nonlinear model – ground truth – and the one predicted with the linear, lifted model – prediction. In this comparison, the point mass is released from an initial position where both cables are slack. The point mass bounces as one of the cables becomes taut and immediately goes slack. With some damping at the cable, the mass converges to an equilibrium position after a series of bounces, as shown with a black solid line in Fig. 6-(a). This nonlinear dynamic response is predicted almost perfectly with the linear model in the lifted space shown in red. Figure 6-(b)-(e) show the trajectories of the individual state variables. The predicted trajectories shown in red agree with the ground truth black lines even after many bounces. Comparing to the prior work on multi-cable manipulation [30], where a data-driven method based on DMD was used, the proposed method shows an order-of-magnitude more accurate prediction, although the system structure is different.

This linear model incorporates the four dynamic equations into a single unified representation. Again, the results are obtained directly from the governing equations, and the model is valid globally.

## VI. Discussion: Applicability of the Koopman Operator to Heterogeneous Dynamical Systems

The current work addresses key conditions in applying the composition operator method to those heterogenous dynamics problems. To find such conditions, the current work took a simple, straightforward approach to converting function transformation equations to time-evolution dynamic equations based on composition operator theory. The conversion hinges on the conditions given by (16) in Proposition 2 and by (26) in Proposition 3. These conditions are also the key to examining the applicability of the Koopman Operator to a class of challenging robotics problems. If one uses the Koopman Operator for merely linearizing a nonlinear dynamical system, where the state transition function $F(x)$ is smooth and continuous, these conditions, (16) and (26), are satisfied for most observable functions, $g_k \in \mathcal{H}$. However, this is not automatically met for heterogeneous dynamical systems, including hybrid and switched systems.

In the proof of Proposition 2, the compositional function $\varphi_j \circ F$ was expanded to $\varphi_j \circ F = \sum_{k=1}^{\infty} \langle \varphi_j \circ F, g_k \rangle g_k$. Through this expansion, observables $g_1, g_2, \ldots$ were pulled out. This is the key mechanism for creating a new extended state space with state variables $\chi = [g_1, g_2, \cdots]^T$, leading to the formation of matrix $A$. In Proposition 2, it was assumed that $\{g_1, g_2, \ldots\}$ forms an orthonormal basis, but this condition was relaxed in Proposition 3, where $g_1, g_2, \ldots$ are an independent and complete set of basis functions.

These existence conditions are dependent on the selection of observables as well as on the properties of state transition function $F(x)$. The first example discussed above is a univariate heterogenous dynamical system with discontinuities in state transition. For this system, properties of the compositional function, $\varphi_i \circ F$, can be examined analytically. Although discontinuous transitions are involved, a Fourier series exists for an arbitrary $j$ that is finite. This allows us to expand $\varphi_i \circ F$ to $\sum_{k=1}^{\infty} \langle \varphi_i \circ F, g_k \rangle g_k$ and obtain the time-evolution dynamic equation. This does not necessarily mean $\lim_{i \to \infty} \varphi_i \circ F \in \mathcal{H}$ because the compositional function may have an infinite number of discontinuities in a finite interval, which is against the conditions for the existence of Fourier series expansion. Nonetheless, for *any* finite order $i < \infty$, $\varphi_i \circ F \in \mathcal{H}$. This property supports the high accuracy prediction that is achieved with a large number of observables, as confirmed in Fig.3-E.

For practical implementation, the infinite-dimensional observable vector $\chi$ must be truncated. In doing so, the condition:

$$\varphi_i \circ F \in \mathcal{H}, \quad 1 \leq \forall i < \infty \tag{35}$$

is a useful property that underpins high-accuracy, high-dimensional truncation. Relaxing the conditions for the existence of a time-evolution $A$ matrix to the above condition, $\forall i < \infty$, allows us to apply the composition operator method to broader systems, including hybrid and switched systems with discontinuities. In fact, the prediction based on the $A_m$ matrix under the conditions (35) showed an extremely high accuracy over a long period of time, as demonstrated in Fig.3.

In general, it is difficult to analytically examine $\varphi_i \circ F \in \mathcal{H}$ for more general, multivariate systems with complex heterogeneous dynamics. However, there are a few methods for numerically examining the conditions [27],[31].

- Compute $J_N = \sum_{k=1}^{N} |\langle \varphi_i \circ F, \varphi_k \rangle|^2 \leq \|\varphi_i \circ F\|$ and examine whether $J_N$ approaches $\|\varphi_i \circ F\|$ as $N$ tends to infinity.
- Compute $I_N = \sum_{k=1}^{N} \langle \varphi_i \circ F, \varphi_k \rangle \varphi_k$ and examine whether $I_N$ approaches $\varphi_i \circ F$ as $N$ tends to infinity.

Using these metrics, we can examine the conditions.

The second example, the multi-cable suspension problem, is a multivariate, heterogenous dynamical system described with 4 sets of dynamic equations. This system does not have any discontinuity in state transition; both position and velocity of the mass are continuous, as long as the stiffness of the cables is finite. The 4 dynamic equations have been combined and transformed to a single linear dynamic equation, which has been shown extremely accurate. The linear model can predict the mass trajectory with high accuracy even after many bounces. The high accuracy modeling was achieved through direct encoding of the governing equations, rather than curve fitting to numerical data.

As the extensile cable stiffness increases to infinite stiffness, the velocity of the mass tends to change discontinuously when bouncing. To deal with discontinuous state transitions, the method based on Proposition 4 is required, as in the case of the univariate example. This entails truncation of the lifted space dimension, which leads to approximation. However, truncation is necessary for practical implementation. Therefore, this does not degrade the usefulness of the composition operator approach and its applicability to heterogenous dynamical system with discontinuous state transition.

Overall, the proposed composite operator method for global, unified representation of heterogenous robot dynamics is applicable to various robotics problems, including hybrid and switched systems with discontinuities.

A few caveats must be stated, however. First, a large number of observables are required for dealing with discontinuous state transitions. This can be alleviated by using observable functions that are locally tunable. RBFs, for example, can be allocated densely to a region near the discontinuity [31]. See Appendix B. Second, the Direct Encoding method is difficult to apply to a high-order nonlinear system due to the multivariate inner product integration. As the number of independent state variables, $n$, exceeds 6 or 7, it is difficult to perform the computation of inner products. This necessitates an effective computational algorithm.

## VII. CONCLUSION

The current work demonstrates that complex robot dynamics given by heterogenous governing equations can be represented with a single, unified linear model in a lifted space. Although the original dynamics are highly nonlinear, segmented, and heterogeneous, the lifted linear model is valid globally. Although the system experiences a multitude of jumps and transitions to diverse dynamic behaviors, the prediction accuracy remains high. In robotics, heterogeneous dynamics are common, due to contact-noncontact transition, non-holonomic constraints, and other discrete transitions among diverse dynamics. The composition operator approach presented in the current work can be a promising methodology in dealing with those heterogeneous dynamics.

In the current work, only autonomous systems having no exogenous input are considered. Extending the original Koopman Operator theory to general control systems is a critically important challenge that has been addressed by several authors and is still an active research area in the field [17], [19], [32], [33].

The theory established for autonomous systems, including the current work, can be extended without any significant change to some class of non-autonomous systems, if they meet the following conditions.
- Control $u_t$ comes into a system as a linear term: $x_{t+1} = F(x_t) + Bu_t$. In this case the effect of the input term $Bu_t$ can be separated, and the matrix $B$ can be determined by solving a least squares estimate problem [8], [17], [33] ; or
- If the control is a feedback regulator, $u_t = h(x_t)$, then the problem reduces to an autonomous system by embedding the control law into the system [17], [19].

These are simple cases to which the existing methods for autonomous systems can be applied. For other general cases, the Koopman methods, including the current work, must be modified substantially. An elegant solution to this problem is to represent a control system as a bilinear system, where inputs are multiplied with functions of state [34]. The resultant system is not linear, but the Koopman operator can be applied more rigorously. More research effort is needed to extend the current work to those general control cases.

Although the general non-autonomous system's problem has not yet been solved, it has been reported that the Koopman-based lifting linearization is effective for controlling a class of nonlinear systems. The multi-cable system described in Fig.4 has been controlled with MPC [30]. The highly nonlinear, heterogenous dynamics are represented with a single linear state equation, which allows for efficient real-time MPC computation. Unlike nonlinear MPC, the linear MPC is a convex-optimization computation with no local minima problem. Reference [35] reported that the computation time reduced approximately 100 times compared to nonlinear MPC.

Finding an effective set of observables remains a challenging problem faced by the Koopman operator research community [36]. In particular, the proposed method entails effective observables that meet the critical condition $\varphi_i \circ F \in \mathcal{H}$, which depends on both the properties of the observables and the heterogenous dynamics of the system. Effective algorithms must be developed for searching for observables that can approximate the system with fewer variables and that can keep the compositional function in a Hilbert space despite the heterogenous nature of the dynamics.

Despite the limitations to the current method, the proposed method is a promising approach to dealing with challenging robotics problems delineated as heterogeneous dynamics. The proposed method has the potential to provide the robotics community with a powerful tool: global, unified representation of otherwise complex heterogenous robot dynamics, which is also amenable to analysis and synthesis.

## APPENDIX

APPENDIX A: Computation of matrices $C$ and $C^{-1}$ in (25)

We use the exponential trigonometric functions: $\varphi_k = \exp(2\pi i k x)$. The observables are related to these as: $g_1 = 1 = \varphi_0$, $g_{2n} = \cos 2\pi n x = \frac{1}{2}(\varphi_n + \varphi_{-n})$, $g_{2n+1} = \sin 2\pi n x = -\frac{i}{2}(\varphi_n - \varphi_{-n})$, $n \in \mathbb{N}$. We can find the $C$ matrix as:

$$\begin{bmatrix} g_1 \\ g_2 \\ g_3 \\ \vdots \\ \vdots \\ \vdots \end{bmatrix} = \begin{bmatrix} 1 & 0 & 0 & 0 & 0 & \cdots \\ 0 & .5 & .5 & 0 & 0 & \cdots \\ 0 & -.5i & .5i & 0 & 0 & \cdots \\ 0 & 0 & 0 & .5 & .5 & \cdots \\ 0 & 0 & 0 & -.5i & .5i & \cdots \\ \vdots & \vdots & \vdots & \vdots & \vdots & \ddots \end{bmatrix} \begin{bmatrix} \varphi_0 \\ \varphi_1 \\ \varphi_{-1} \\ \varphi_2 \\ \varphi_{-2} \\ \vdots \end{bmatrix}$$

Also, from $\varphi_n = \exp(2\pi i n x) = \cos 2\pi n x + i \sin 2\pi n x = g_{2n} + ig_{2n+1}$, and $\varphi_{-n} = \exp(-2\pi i n x) = \cos 2\pi n x - i \sin 2\pi n x = g_{2n} - ig_{2n+1}$, $C^{-1}$ can be found.

$$C^{-1} = \begin{bmatrix} 1 & 0 & 0 & 0 & 0 & \cdots \\ 0 & 1 & i & 0 & 0 & \cdots \\ 0 & 1 & -i & 0 & 0 & \cdots \\ 0 & 0 & 0 & 1 & i & \cdots \\ 0 & 0 & 0 & 1 & -i & \cdots \\ \vdots & \vdots & \vdots & \vdots & \vdots & \ddots \end{bmatrix}$$

APPENDIX B: Tuning of Radial Basis Functions

The RBFs must be tuned properly so that a finite number of RBFs can approximate the system effectively. The most critical is the locations of center points $c_k$. A clustering technique was applied to find effective locations of the center points.

The overall prediction accuracy of the linear model, $\chi(t+1) = A\chi(t)$, was evaluated by comparing the predicted trajectories against the ones computed from the nonlinear model, $x_{t+1} = F(x_t)$. In doing so, several ground truth trajectories were created. Using these as sample points, we can find where RBFs should be placed densely or sparsely. We applied the $k$ means clustering algorithm with an improved initial setting based on the $k$ mean ++ algorithm. Fig. 7 shows the optimized locations of the center points. High-density center points can be found in the vicinity of the locations where one of the cables switches between taut and slack conditions. We specified the number of clusters to be 1,500 for placing 1,500 RBFs.

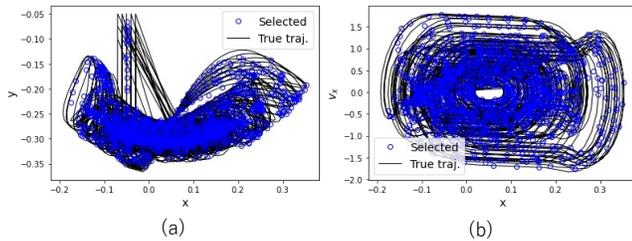

Fig. 7 Center points of RBFs are determined using k mean++ clustering algorithm applied to sample trajectories

## REFERENCES

1. M. H. Raibert, Legged Robots That Balance. Cambridge, MA: MIT Press, (1986).
2. J.E. Colgate, N. Hogan, Robust Control of Dynamically Interacting Systems, International Journal of Control, 48, 65-85, (1988).
3. A. Hassibi, S. Boyd, and J. How, A class of Lyapunov Functionals for Analyzing Hybrid Dynamical Systems," in Proceedings of the 1999 American Control Conference, 4, 2455–2460 (1999).
4. S. Chen, M. Fazlyab, M. Morari, G. Pappas, V. Preciado, Learning Lyapunov Functions for Hybrid Systems, in the Proceedings of the 2021 55-th Annual Conference on Information Sciences and Systems, (2020), DOI:10.1109/CISS50987.2021.9400289.
5. W. Znegui, H. Gritli and S. Belghith, An Enhanced Poincaré Map Expression for the Passive Dynamic Walking of the Compass-Gait Biped Robot, 2020 17th International Multi-Conference on Systems, Signals & Devices, 720-726, (2020), doi: 10.1109/SSD49366.2020.9364265.
6. E. Summers, A. Chakraborty, W. Tan, U. Topcu, P. Seiler, G. Balas, and A. Packard, Quantitative Local l2-Gain and Reachability Analysis for Nonlinear Systems, International Journal of Robust and Nonlinear Control, 23, 10, 1115–1135, (2013).
7. J. Maidens and M. Arcak, Reachability Analysis of Nonlinear Systems Using Matrix Measures," IEEE Transactions on Automatic Control, 60, 1, 265–270, (2015).
8. H. Asada, F. Sotiropoulos, Dual Faceted Linearization of Nonlinear Dynamical Systems Based on Physical Modeling Theory, ASME Journal of Dynamic Systems, Measurement, and Control, 141-2, (2019), https://doi.org/10.1115/1.4041448.
9. J.E. Littlewood, On Inequalities in the Theory of Functions, Proceedings of London Mathematics Society, 23, 481-519, (1925).
10. B. O. Koopman, Hamiltonian Systems and Transformation in Hilbert Space, Proceedings of the National Academy of Sciences, 17-5, 315–318, (1931).
11. C. W. Rowley, I. Mezic, S. Bagheri, P. Schlatter, D.S. Henningson, Spectral Analysis of Nonlinear Flows, Journal of Fluid Mechanics, 641, 1-13, (2009), doi:10.1017/S0022112009992059.
12. Y. Susuki and I. Mezic, "Nonlinear Koopman Modes and Coherency Identification of Coupled Swing Dynamics, IEEE Transactions on Power Systems, 26-4, 1894-1904, (2011), doi: 10.1109/TPWRS.2010.2103369.
13. I. Mezic, Analysis of Fluid Flows via Spectral Properties of the Koopman Operator, Annual Review of Fluid Mechanics, 45, 357-378, (2013), https://doi.org/10.1146/annurev-fluid-011212-140652.
14. M. Williams, I. G. Kevrekidis, C. W. Rowley, A Data–Driven Approximation of the Koopman Operator: Extending Dynamic Mode Decomposition, Journal of Nonlinear Science, 25, 1307-1346, (2015), DOI 10.1007/s00332-015-9258-5.
15. J. H. Tu, C. W. Rowley, D. M. Luchtenburg, S. L. Brunton, J. N. Kutz. On Dynamic Mode Decomposition: Theory and Applications. Journal of Computational Dynamics, 1-2, 391-421. (2014), doi: 10.3934/jcd.2014.1.391.
16. H. Arbabi, I. Mezic, Ergodic Theory, Dynamic Mode Decomposition, and Computation of Spectral Properties of the Koopman Operator, SIAM Journal of Applied Dynamical Systems, 16 – 4, (2017), https://doi.org/10.1137/17M1125236.
17. J. L. Proctor, S. L. Brunton, and J. N. Kutz, Dynamic Mode Decomposition with Control, SIAM Journal of Applied Dynamical Systems, 15-1, 142-161, (2016).
18. M. Korda, I. Mezic, Linear Predictors for Nonlinear Dynamical Systems: Koopman Operator Meets Model Predictive Control, Automatica, 93, 149-160, (2018), https://doi.org/10.1016/j.automatica.2018.03.046.
19. S. L. Brunton1, B. W. Brunton, J. L. Proctor, J. N. Kutz, Koopman Invariant Subspaces and Finite Linear Representations of Nonlinear Dynamical Systems for Control, PLoS ONE, (2016), https://doi.org/10.1371/journal.pone.0150171.
20. I. U. Haq, T. Iwata, Y. Kawahara, Dynamic Mode Decomposition via Convolutional Autoencoders for Dynamics Modeling in Videos, Computer Vision and Image Understanding, 216, (2022).
21. Y. Xiao, Z. Tang, X. Xu, L. Qian, A Koopman-based Deep Convolutional Network for Modeling Latent Dynamics from Pixels, arXiv:2102.10205v2 [eess.SY] 27 Jul (2021).
22. I. Abraham, T. Murphey, Active Learning of Dynamics for Data-Driven Control Using Koopman Operators, IEEE Trans. on Robotics, 35-5, 1071-1083, (2019).
23. D. Bruder, X. Fu, R. B. Gillespie, C. D. Remy, R.Vasudevan, Koopman-Based Control of a Soft Continuum Manipulator Under Variable Loading Conditions, IEEE Robotics and Automation Letters, 6 - 4, (2021).
24. M. L. Castano, A. Hess, G. Mamakoukas, T. Gao, T. Murphey, X. Tan, Control-oriented Modeling of Soft Robotic Swimmer with Koopman Operators, Proc. of the 2020 IEEE/ASME International Conference on Advanced Intelligent Mechatronics, (2020).
25. A. Broad, I. Abraham, T. Murphy, B. Argall, Data-Driven Koopman Operators for Model-Based Shared Control of Human-Machine Systems, International Journal of Robotics Research, 39-9, 1178 – 1195, (2020).
26. P. Schmid, "Dynamical Mode Decomposition of Numerical and Experimental Data", Journal of Fluid Mechanics, Vol.656, pp. 5 – 28, (2010) , DOI:10.1017/S0022112010001217.
27. J. Shapiro, Composition Operators and Classical Function Theory, Springer-Verlag, (1993).
28. A. Mauroy, Y. Susuki, and I. Mezi ́c, "Introduction to the koopman operator in dynamical systems and control theory," The Koopman Operator in Systems and Control, Chapter 1, pp. 3–33, Springer, 2020.
29. A. V. Oppenheim, A. S. Willsky, Signals and Systems, 2nd ed. Prentice Hall, (1983).
30. J. Ng, H. H. Asada, Model Predictive Control and Transfer Learning of Hybrid Systems Using Lifting Linearization Applied to Cable Suspension Systems," IEEE Robotics and Automation Letters, 7 - 2, 682–689, (2022).
31. M. D. Buhmann, Radial Basis Functions: Theory and Implementation, Cambridge University Press, (2003).
32. N. Selby, H.H. Asada, Learning of Causal Observable Functions for Koopman-DFL Lifting Linearization of Nonlinear Controlled Systems and Its Application to Excavation Automation, IEEE Robotics and Automation Letters, 6-4, 6297-6304, (2021), DOI: 10.1109/LRA.2021.3092256.
33. N. Selby, F. Sotiropoulos, H.H. Asada, Physics-Based Causal Lifting Linearization of Nonlinear Control Systems Underpinned by the Koopman Operator, arXiv:2108.10980v1 [eess.SY] 24 Aug (2021).
34. D. Bruder, X. Fu, R. Vasudevan, Advantages of Bilinear Koopman Realizations for the Modeling and Control of Systems With Unknown Dynamics, IEEE Robotics and Automation Letters, 6 – 3, 4369 – 4376, (2021).
35. Y. Igarashi, M. Yamakita, J. Ng, and H. Asada, "MPC Performances for Nonlinear Systems Using Several Linearization Models", Proc. 2020 American Control Conference, doi:10.23919/ACC45564.2020.9147306, Denver, CO, 2020.
36. E. Yeung, S. Kundu, N. Hodas, Learning Deep Neural Network Representation for Koopman Operators of Nonlinear Dynamical Systems, arXiv:1708.06850v2 [cs.LG] 17 Nov 2017, (2017).

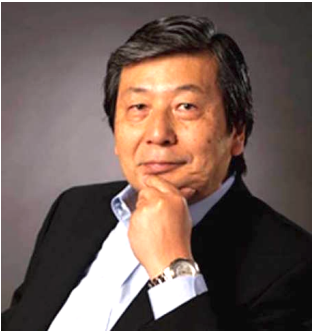

**H. Harry Asada** is Ford Professor of Mechanical Engineering and Director of the Brit and Alex d'Arbeloff Laboratory for Information Systems and Technology in the Department of Mechanical Engineering, Massachusetts Institute of Technology (MIT), Cambridge, MA. He specializes in robotics, system dynamics and control, and biomedical engineering. His current research includes Koopman operator theory, assistive robotics for eldercare, supernumerary robotic limbs, and multi-cable manipulation. He received Best Paper Awards at the IEEE International Conference on Robotics and Automation (ICRA) in 1993, 1997, 1999, and 2010, the best application paper award at the 2017 IEEE/RSJ International Conference on Intelligent Robotics and Systems (IROS), the O. Hugo Schuck Best Paper Award from the American Control Council in 1985, and other 6 best paper awards of major journals and conferences. He was the recipient of the Henry Paynter Outstanding Researcher Award from ASME Dynamic Systems and Control in 1998. More recently he received the 2011 Rufus Oldenburger Medal from ASME, and Ruth and Joel Spira Award for Distinguished Teaching from School of Engineering, MIT. He is a fellow of IEEE and ASME.